\documentclass[final,12pt]{clear2022}

\title[Amortized Causal Discovery]{Amortized Causal Discovery: \\ Learning to Infer Causal Graphs from Time-Series Data}
\usepackage{times}

\usepackage[utf8]{inputenc}
\usepackage{hyperref}
\usepackage{url}

\usepackage{microtype}
\usepackage{appendix}
\usepackage{caption}
\usepackage{graphicx, color}
\usepackage{booktabs}       %
\usepackage{amsfonts}       %
\usepackage{nicefrac}       %
\usepackage{microtype}      %
\usepackage{multirow, multicol}
\usepackage{ragged2e}
\usepackage{bm, bbm}
\usepackage{booktabs} %
\usepackage{enumitem}
\usepackage{tikz}
\usepackage{wrapfig}
\usetikzlibrary{positioning}
\usepackage{lipsum}

\usepackage[utf8]{inputenc} %
\usepackage[T1]{fontenc}    %
\usepackage{hyperref}       %
\usepackage{url}            %
\usepackage{booktabs}       %
\usepackage{amsfonts}       %
\usepackage{nicefrac}       %
\usepackage{microtype}      %
\usepackage{mathtools}
\usepackage{floatrow} 
\usepackage{algorithm}
\usepackage{algorithmic}

\usepackage{microtype}
\usepackage{graphicx}
\usepackage{booktabs} %
\usepackage{makecell}
\usepackage{hyperref}

\usepackage{natbib}

\usepackage{bm}
\usepackage{amsmath}
\usepackage[capitalise]{cleveref}

\usepackage{dsfont}
\usepackage{enumitem}
\usepackage{floatrow}
\usepackage{wrapfig}
\usepackage{amssymb}
\usepackage{caption}
\usepackage{adjustbox}

\newcommand{\xvec}{\bm x}
\newcommand{\gvec}{\bm g}
\newcommand{\hvec}{\bm h}
\newcommand{\wvec}{\bm w}
\newcommand{\psivec}{\bm \psi}
\newcommand{\zvec}{\bm z}
\newcommand{\muvec}{\bm \mu}

\newcommand{\XMat}{\bm X}
\newcommand{\zij}{\bm z_{ij}}
\newcommand{\xvect}{\xvec^t}

\newcommand{\expected}{\mathds{E}}
\newcommand{\KL}{\textrm{KL}}
\newcommand{\elbo}{\mathcal{L}}
\newcommand{\qphi}{q_{\phi}(\zvec | \xvec)}
\newcommand{\qz}{q(\zvec)}

\newcommand{\qphizij}{q_{\phi}(\zij |\xvec)}

\newcommand{\ptheta}{p_{\theta}(\xvec | \zvec)}

\newcommand{\prior}{p(\zvec)}

\newcommand{\softmax}[1]{\texttt{Softmax}\,(#1)}

\newcommand{\Xtrain}{\XMat_{\text{train}}}
\newcommand{\xtest}{\xvec_{\text{test}}}

\clearauthor{%
 \Name{Sindy Löwe} \Email{loewe.sindy@gmail.com}\\
 \addr UvA-Bosch Delta Lab, University of Amsterdam
 \AND
 \Name{David Madras} \Email{madras@cs.toronto.edu}\\
 \addr University of Toronto, Vector Institute
  \AND
 \Name{Richard Zemel} \Email{zemel@cs.toronto.edu}\\
 \addr University of Toronto, Vector Institute
  \AND
 \Name{Max Welling} \Email{welling.max@gmail.com}\\
 \addr UvA-Bosch Delta Lab, University of Amsterdam
}

\begin{document}

\maketitle

\begin{abstract}%
  On time-series data, most causal discovery methods fit a new model whenever they encounter samples from a new underlying causal graph. However, these samples often share relevant information which is lost when following this approach. Specifically, different samples may share the dynamics which describe the effects of their causal relations. We propose Amortized Causal Discovery, a novel framework that leverages such shared dynamics to learn to infer causal relations from time-series data. This enables us to train a single, amortized model that infers causal relations across samples with different underlying causal graphs, and thus leverages the shared dynamics information. We demonstrate experimentally that this approach, implemented as a variational model, leads to significant improvements in causal discovery performance, and show how it can be extended to perform well under added noise and hidden confounding.
\end{abstract}


\begin{figure*}
    \centering
    \includegraphics[width=0.85\linewidth]{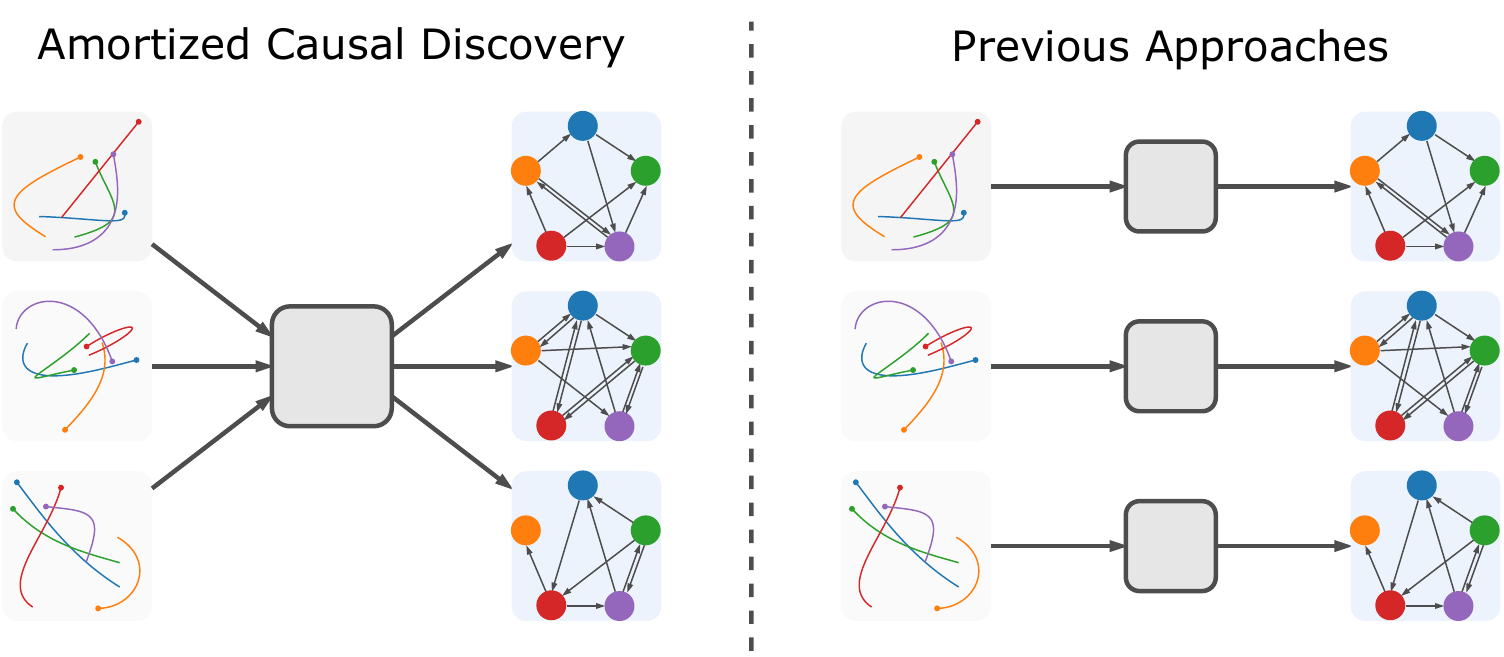}
    \caption{Amortized Causal Discovery. We propose to train a single model that predicts causal relations across samples with different underlying causal graphs but shared dynamics (\cref{eq:dynamics}). This allows us to generalize across samples and to improve our performance with additional training data. In contrast, previous approaches (\cref{sec:background}) fit a new model for every sample with a different underlying causal graph.}
    \label{fig:eye_catcher}
\end{figure*}

\section{Introduction}
\label{sec:intro}

Inferring causal relations in observational time-series is central to many fields of scientific inquiry \citep{spirtes2000causation,berzuini2012causality}.
Suppose you want to analyze fMRI data, which measures the activity of different brain regions over time -- how can you infer the (causal) influence of one brain region on another?
This question is addressed by the field of \textit{causal discovery} \citep{glymour2019review}. Methods within this field allow us to infer causal relations from observational data -- when interventions (e.g. randomized trials) are infeasible, unethical or too expensive.

In time-series, the assumption that causes temporally precede their effects enables us to discover causal relations in observational data \citep{peters2017elements}; with approaches relying on conditional independence tests \citep{entner2010causal}, scoring functions \citep{chickering2002optimal}, or deep learning \citep{tank2018neural}. 
All of these methods assume that samples share a single underlying causal graph and refit a new model whenever this assumption does not hold.
However, samples with different underlying causal graphs may share relevant information, such as the dynamics describing the effects of causal relations. 
For instance, we may want to infer synaptic connections (i.e. causal relations) between neurons based on their spiking behavior, from a set of recordings of neuronal firing.
Despite recording different populations of neurons with different wiring, the dynamics of how neurons connected by synapses influence one another may stay the same.
This principle occurs in a number of areas:
fMRI test subjects may have varying brain connectivity but the same underlying neurochemistry; social networks may have differing structure but comparable interpersonal relationships; different stocks may relate differently to one another but obey similar market forces.
Despite a range of relevant applications, inferring causal relations across samples with different underlying causal graphs is as of yet largely unexplored.

In this paper, we propose a novel causal discovery framework for time-series that embraces this aspect: Amortized Causal Discovery (\cref{fig:eye_catcher}). 
In this framework, we learn to infer causal relations across samples with different underlying causal graphs but shared dynamics. 
We achieve this by separating the causal relation prediction from the modeling of their dynamics:
an amortized encoder predicts the edges in the causal graph, and a decoder models the dynamics of the system under the predicted causal relations.
This setup allows us to pool statistical strength across samples and to achieve significant improvements in performance with additional training data. 
It also allows us to infer causal relations in previously unseen samples without refitting our model.
Additionally, we show that Amortized Causal Discovery can improve robustness under hidden confounding by modeling the unobserved variables with the amortized encoder.
Our contributions are as follows:
\begin{itemize} %
    \item We formalize Amortized Causal Discovery (ACD), a novel framework for causal discovery in time-series, in which we learn to infer causal relations from samples with different underlying causal graphs but shared dynamics (\cref{eq:dynamics}).
    \item We propose a variational model for ACD, applicable to multi-variate, non-linear data.
    \item We present experiments demonstrating the effectiveness of this model on a range of causal discovery datasets, in the fully observed setting, with added noise, and under hidden confounding.
\end{itemize}

\section{Background: Granger Causality} \label{sec:background}

Granger causality \citep{granger1969investigating} is one of the most commonly used approaches to infer causal relations from observational time-series data.
Its central assumption is that causes precede their effects: if the prediction of the future of time-series $Y$ can be improved by knowing past elements of time-series $X$, then $X$ ``Granger causes'' $Y$.
Originally, Granger causality was defined for linear relations; we follow the more recent definition of \citet{tank2018neural} for non-linear Granger causality: 

\begin{definition}{\textit{Non-Linear Granger Causality}:}\label{def:nonlinear_granger} %
Given $N$ stationary time-series $\xvec = \{\xvec_1, ... \xvec_N\}$ across time-steps $t = \{1,...,T \}$ and a non-linear autoregressive function $g_j$, such that
\begin{align} \label{eq:nodes}
    \xvec^{t+1}_j = g_j(\xvec^{\leq t}_1, ..., \xvec^{\leq t}_N) + \bm \varepsilon^{t+1}_j  ~~~, 
\end{align} 
where $\xvec^{\leq t}_j = (..., \xvec^{t-1}_j, \xvec^{t}_j)$ denotes the present and past of series $j$ and $\bm \varepsilon^{t+1}_j$ represents independent noise. 
In this setup, time-series $i$ Granger causes $j$, if $g_j$ depends on $\xvec^{\leq t}_i$, i.e. if \\
$\exists~ \xvec^{\prime \leq t}_{i} \neq \xvec^{\leq t}_i : ~~ g_j(\xvec^{\leq t}_1, ..., \xvec^{\prime \leq t}_{i}, ..., \xvec^{\leq t}_N) \neq g_j(\xvec^{\leq t}_1, ..., \xvec^{\leq t}_i, ... \xvec^{\leq t}_N)$.
\end{definition}

Granger causal relations are equivalent to causal relations in the underlying directed acyclic graph (DAG) if all relevant variables are observed and no instantaneous\footnote{connections between two variables at the same time step} connections exist \cite[Theorem 10.1]{peters2013causal,peters2017elements}. %

Many methods for Granger causal discovery, including vector autoregressive  \citep{JMLR:v11:hyvarinen10a} and more recent deep learning-based approaches \citep{tank2018neural,khanna2019economy,wu2020discovering}, can be encapsulated by a particular framework:
\begin{enumerate} %
    \item Define a function $f_\theta$ (an MLP in \citet{tank2018neural}, a linear model in \citet{JMLR:v11:hyvarinen10a}), which learns to predict the next time-step of a given test sequence $\xtest$.
    \item Fit $f_\theta$ to $\xtest$ by minimizing some loss $\mathcal{L}$: $\theta_\star = \textrm{argmin}_{\theta} \ \mathcal{L}(\xtest, f_\theta)$.
    \item Apply some fixed function $h$ (e.g. thresholding) to the learned parameters to produce the Granger causal graph estimate for $\xtest$: 
    $\hat{\mathcal{G}}_{\xtest} = h(\theta_\star)$.
    For instance, \citet{tank2018neural} infer the Granger causal relations through examination of the weights $\theta_\star$: if all outgoing weights $\wvec_{ij}$ between time-series $i$ and $j$ are zero, then $i$ does not Granger-cause $j$. 
\end{enumerate}
The shortcoming of this approach is that, when we have $S$ samples $\xvec_1, \dots, \xvec_S$ with different underlying causal graphs, the parameters $\theta$ must be optimized separately for each of them.
As a result, methods within this framework cannot take advantage of the information that might be shared between samples.
This motivates us to question: can we amortize this process?

\section{Amortized Causal Discovery} \label{sec:method}

We propose Amortized Causal Discovery (ACD), a framework in which we learn to infer causal relations across samples with different underlying causal graphs but shared dynamics.
To illustrate, we return to the example from \cref{sec:intro}: suppose you want to infer synaptic connections (i.e. causal relations) between neurons based on their spiking behavior. 
You are given a set of $S$ recordings (i.e. samples), each containing $N$ time-series representing the firing of $N$ individual neurons. Even though you might record across different populations of neurons with different wiring, the dynamics of how neurons connected by synapses influence one another stays the same. ACD takes advantage of such shared dynamics to improve the prediction of causal relations.
Given a training set $\Xtrain$ and test sequence $\xtest$, it can be summarized as follows:
\begin{enumerate} %
    \item Define an encoding function $f_\phi$ which learns to infer Granger causal relations of any sample $\xvec_i$ in $\Xtrain$. Define a decoding function $f_\theta$ which learns to predict the next time-step of the samples under the inferred causal relations.
    \item Fit $f_\phi$ and $f_\theta$ to $\Xtrain$ by minimizing some %
    loss $\mathcal{L}$: $f_{\phi_\star}, f_{\theta_\star} = \textrm{argmin}_{f_\phi, f_\theta} \ \mathcal{L}(\Xtrain, f_\phi, f_\theta)$.
    \item For a test sequence $\xtest$, simply output the Granger causal graph estimate $\hat{\mathcal{G}}_{\xtest}$: \\ $\hat{\mathcal{G}}_{\xtest} = f_{\phi_\star}(\xtest)$.
\end{enumerate}
By dividing the model into two parts, an encoder and a decoder, ACD can use the \textit{activations} of $f_{\phi_\star}$ to infer causal structure. This increases the flexibility of our approach greatly compared to methods that use the learned \textit{weights} $\theta_\star$ such as the prior Granger causal discovery methods described in \cref{sec:background}.
In this section, we describe our framework in more detail, and provide a probabilistic implementation thereof. We also extend our approach to model hidden confounders.

\paragraph{Preliminaries}
We begin with a dataset $\XMat = \{\xvec_s\}_{s=1}^S$ of $S$ samples, where each sample $\xvec_s$ consists of $N$ stationary time-series $\xvec_s = \{\xvec_{s,1}, \dots, \xvec_{s,N} \}$ across time-steps $t = \{1,...,T \}$. 
We denote the $t$-th time-step of the $i$-th time-series of $\xvec_s$ as $\xvec_{s,i}^t$ (sometimes omitting $s$ for brevity).
We assume there is an associated directed acyclic graph $\mathcal{G}^{1:T}_s = \{\mathcal{V}^{1:T}_s, \mathcal{E}^{1:T}_s\}$ underlying the generative process of each sample. 
This is a structural causal model (SCM) \citep{pearl2009causality}. Its endogenous (observed) variables are vertices $v_{s,i}^t \in \mathcal{V}_{s}^{1:T}$ for each time-series $i$ and each time-step $t$. Every set of incoming edges to an endogenous variable defines inputs to a deterministic function $g_{s, i}^t$ which determines that variable's value\footnote{The SCM also includes an exogenous (unobserved), independently-sampled error variable $\varepsilon_{v}$ as a parent of each vertex $v$, which we do not model and thus leave out for brevity.}.
The edges are defined by ordered pairs of vertices $\mathcal{E}^{1:T}_s = \{ (v_{s,i}^t, v_{s,j}^{t'}) \}$, which we make two assumptions about:
\begin{enumerate} %
    \item No edges are instantaneous ($t = t'$) or go back in time. Thus, $t < t'$ for all edges.
    \item Edges are invariant to time. 
    Thus, if $(v_{s,i}^t, v_{s,j}^{t + k})  \in \mathcal{E}^{1:T}_s$, then $\forall 1 \leq t' \leq T - k:  ( v_{s,i}^{t'}, v_{s,j}^{t' + k}) \in \mathcal{E}^{1:T}_s$.
    The associated structural equations $g_{s, i}^t$ are invariant to time as well, i.e. $g_{s, i}^t = g_{s, i}^{t'} \ \forall t, t'$.
\end{enumerate}
The first assumption states that causes temporally precede their effects and makes causal relations identifiable from observational data, when no hidden confounders are present
\cite[Theorem 10.1]{peters2013causal,peters2017elements}.
The second simplifies modeling: it is a fairly general assumption that allows us to define dynamics that govern all time-steps (\cref{eq:dynamics}). 

Throughout this paper, we are interested in discovering the \textit{summary graph} $\mathcal{G}_s = \{\mathcal{V}_s, \mathcal{E}_s\}$ \citep{peters2017elements}.
It consists of vertices $v_{s,i} \in \mathcal{V}_s$ for each time-series $i$ in sample $s$, and has directed edges whenever they exist in $\mathcal{E}^{1:T}_s$ at any time-step,
i.e. $\mathcal{E}_s = \{ (v_{s,i}, v_{s,j}) \mid \exists t, t': (v_{s,i}^t, v_{s,j}^{t'}) \in \mathcal{E}^{1:T}_s\}$.
Note that while $\mathcal{G}^{1:T}_s$ is acyclic (due to the first assumption above), %
the summary graph $\mathcal{G}_s$ may contain (self-)cycles. %

\paragraph{Amortized Causal Discovery}
The key assumption for Amortized Causal Discovery is that there exists some fixed function $g$ that describes the dynamics of \textit{all} samples $\xvec_s \in \XMat$ given their past observations $\xvec^{\leq t}_s$ and their underlying causal graph $\mathcal{G}_s$:
\begin{align}\label{eq:dynamics}
    \xvec^{t+1}_s = g(\xvec_s^{\leq t}, \mathcal{G}_s) + \bm \varepsilon_s^{t+1} ~~~.
\end{align} 
There are two variables in this data-generating process that we would like to model: the causal graph $\mathcal{G}_s$ that is specific to sample $\xvec_s$, and the dynamics $g$ that are shared across all samples. 
This separation between the causal graph and the dynamics allows us to divide our model accordingly: we introduce an amortized causal discovery encoder $f_\phi$ which learns to infer a causal graph $\mathcal{G}_s$ given the sample $\xvec_s$, and a dynamics decoder $f_\theta$ that learns to approximate $g$:
\begin{align}
    \xvec^{t+1}_s \approx f_\theta(\xvec_s^{\leq t}, f_\phi(\xvec_s)) ~~~.
\end{align}

We formalize Amortized Causal Discovery (ACD) as follows.
Let $\mathds{G}$ be the domain of all possible summary graphs on $\xvec_s$: $\mathcal{G}_s \in \mathds{G}$.
Let $\mathds{X}$ be the domain of any single step, partial or full, observed sequence: $\xvec_{s}^t, \xvec_s^{\leq t}, \xvec_{s} \in \mathds{X}$.
The model consists of two components: a causal discovery encoder $f_\phi: \mathds{X} \rightarrow \mathds{G}$ which infers a causal graph for each input sample, and a decoder $f_\theta:  \mathds{X} \times \mathds{G} \rightarrow \mathds{X}$ which models the dynamics.
This model is optimized with a sample-wise loss $\ell: \mathds{X} \times \mathds{X} \rightarrow \mathds{R}$ which scores how well the decoder models the true dynamics of $\xvec_s$, and a regularization term $r: \mathds{G} \rightarrow \mathds{R}$ on the inferred graphs. For example, this function $r$ may enforce sparsity by penalizing graphs with more edges.
Note, that our formulation of the graph prediction problem is unsupervised: we do \textit{not} have access to the true underlying graph $\mathcal{G}_s$.
Then, given some dataset $\Xtrain$ with $S$ samples, we optimize:
\begin{align}\label{eq:optim}
        f_{\phi_\star}, f_{\theta_\star} &= \textrm{argmin}_{f_\phi, f_\theta} \ \mathcal{L}(\Xtrain, f_\phi, f_\theta)\\
        \textrm{where } \ \mathcal{L}(\Xtrain, \phi, \theta) &= \sum_{s=1}^S \sum_{t=1}^{T-1} \ell(\xvec^{t+1}_s, f_\theta(\xvec_s^{\leq t}, f_\phi(\xvec_s))) + r(f_\phi(\xvec_s)) ~~~.
\end{align}

Once we have completed optimization, we can perform causal graph prediction on any new input test sample $\xtest$ in two ways
 -- we can feed $\xtest$ into the amortized encoder and take its output as the predicted edges (Eq. \ref{eq:enc-optimization}); or we can instantiate our estimate $\hat{\mathcal{G}}_{test} \in \mathds{G}$ which will be our edge predictions, and find the edges which best explain the observed sequence $\xtest$ by minimizing the (learned) decoding loss with respect to $\hat{\mathcal{G}}_{test}$, which we term \emph{Test-Time Adaptation (TTA)} (Eq. \ref{eq:tta}):
\begin{align} 
        \hat{\mathcal{G}}^{\text{Enc}} &= f_{\phi_\star}(\xtest) ~~~; \label{eq:enc-optimization}\\
        \hat{\mathcal{G}}^{\text{TTA}} &= \textrm{argmin}_{\hat{\mathcal{G}}_{test} \in \mathds{G}} \ \mathcal{L}(\xtest, \hat{\mathcal{G}}_{test}, f_{\theta_\star}) ~~~. \label{eq:tta}
\end{align}

By separating the prediction of causal relations from the modeling of their dynamics, ACD yields a number of benefits.
ACD can learn to infer causal relations across samples with different underlying causal graphs, and it can infer causal relations in previously unseen test samples without refitting (\cref{eq:enc-optimization}).
By generalizing across samples, it can improve causal discovery performance with increasing training data size.
We can replace either $f_\phi$ or $f_\theta$ with ground truth annotations, or simulate the outcome of counterfactual causal relations. 
Additionally, ACD can be applied in the standard causal discovery setting, where only a single causal graph underlies all samples, by replacing the amortized encoder $f_\phi$ with an estimated graph $\hat{\mathcal{G}}$ (or distribution over $\mathds{G}$) in \cref{eq:optim}.

\subsection{A Probabilistic Implementation of ACD} \label{sec:variational}
\begin{figure*}
    \centering
    \includegraphics[width=0.99\linewidth]{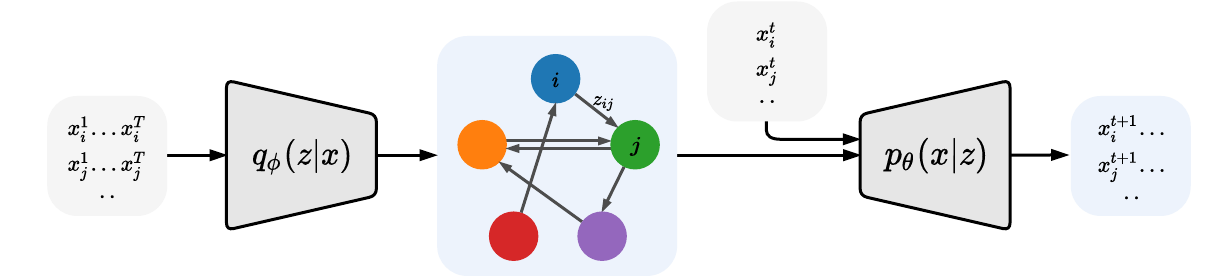}    
    \caption{A Probabilistic Implementation of ACD. An amortized encoder $\qphi$ predicts the causal relations between the input time-series $\xvec$. A decoder $\ptheta$ learns to predict the next time-step of the time-series $\xvec^{t+1}$ given their current values $\xvec^{t}$ and the predicted relations $\zvec$. This separation between causal relation prediction and modeling lets us train the model across samples with different underlying causal graphs but shared dynamics (\cref{eq:dynamics}).}
    \label{fig:our_model}
\end{figure*}
We take a probabilistic approach to ACD and model the functions $f_\phi$ and $f_\theta$ using variational inference (\cref{fig:our_model}).
We amortize the encoder $f_{\phi}$ with an encoding function $\qphi$, which outputs a distribution over $\zvec$ representing the predicted edges $\hat{\mathcal{E}}_{Enc}$ in the causal graph; and we learn a decoder $\ptheta$ which probabilistically models the dynamics of the time-series under the predicted causal relations.
We choose a negative log-likelihood for the decoder loss $\ell$ and a KL-Divergence to a prior distribution over $\mathds{G}$ for the regularizer $r$.
As a result, our loss function $\mathcal{L}$ is a variational lower bound:
\begin{align}\label{eq:elbo}
    \elbo = \expected_{\qphi} [\log \ptheta] - \KL [\qphi || \prior] ~~~.
\end{align}
\paragraph{Encoder}
The encoder $\qphi$ applies a graph neural network $f_{\text{enc},\phi}$ \citep{scarselli2008graph,li2015gated,gilmer2017neural,kipf2016semi} to the input, which propagates information across a fully connected graph $\mathcal{G} = \{\mathcal{V}, \mathcal{E}\}$. This graph includes vertices $v_i \in \mathcal{V}$ for each time-series $i$, and each pair of vertices $(v_i, v_j)$ is connected by an edge.
\begin{align}
    \psivec_{ij} &= f_{\text{enc},\phi} (\xvec)_{ij} \\
    \qphizij &= \softmax{\psivec_{ij} ~/~ \tau}  ~~~.\\
    \intertext{To enable us to backpropagate through the samples of the discrete distribution $\qphizij$, during training, we relax it by adding Gumbel distributed noise $\gvec$ \citep{maddison2016concrete,jang2016categorical}:}
    \zij &\sim \softmax{(\psivec_{ij} + \gvec) ~/~ \tau} ~~~.
\end{align}
The output $\zij$ of the encoder represents the predicted edges $\mathcal{\hat{E}}_{\text{Enc}}$ in the causal graph $\mathcal{\hat{G}}_{\text{Enc}}$.
We consider the possibility that there are $n_{\mathcal{E}}$ different edge-types expressing causal relationships; for instance, inhibitory or excitatory synaptic connections.
Then, more specifically, $z_{ij,e}=1$ expresses that there is a directed edge of type $e$ from time-series $i$ to $j$, where $e \in \{1, \dots, n_{\mathcal{E}}\}$.

\paragraph{Decoder}
The decoder $\ptheta$ models the dynamics of the time-series under the predicted causal relations.
It uses both the predicted causal relations $\zij$ and the feature vectors of the time-series at the current time-step $t$, $\xvec^t = \{\xvect_1, ... \xvect_N\}$ as its input. First, it propagates information along the predicted edges by applying a neural network $f_e$, using the zero function for $f_0$:
\begin{align} \label{eq:edges}
    \hvec^t_{ij} &= \sum_{e>0} z_{ij,e} f_e([\xvec_i^t, \xvec_j^t]) ~~~.
\end{align}
Then, the decoder accumulates the incoming messages to each node and applies a neural network $f_v$ to predict the change between the current and the next time-step:
\begin{align}
    \muvec_j^{t+1} &= \xvect_j + f_v \left(\left[\sum_{i \neq j} \hvec^t_{ij}, \xvect_j \right]\right) \\
    p_{\theta}(\xvec^{t+1}_j | \xvec^t, \zvec) &= \mathcal{N}(\muvec_j^{t+1}, \sigma^2 \mathbb{I}) ~~~.
\end{align}%
In other words, the decoder predicts $\Delta \hat{\xvec}^t$, which is added to the current value of the time-series to yield the prediction for the next time-step $\hat{\xvec}^{t+1} = \xvec^t + \Delta \hat{\xvec}^t$.
Note that this approach assumes the existence of self-edges: that is, the value of $\xvect_j$ always affects the value of $\xvec^{t + 1}_j$.

\paragraph{Prediction of Causal Relations}
In order to align our model with the philosophy of Granger Causality, we include a ``no edge''-type edge function: If the encoder predicts the ``no edge''-type edge $e=0$ by setting $z_{ij,0} = 1$, the decoder uses the zero function and no information is propagated from time-series $i$ to $j$ (\cref{eq:edges}).
Due to this, time-series $i$ will Granger cause the decoder-predicted time-series $j$ only when the edge is predicted to exist (see \cref{app:claim-gc}). 
Hence, by the same logic that justifies prior Granger causal work (\cref{sec:background}), we expect the predicted edges to correspond to Granger causal relations.
Finally, since we assume no hidden confounders and no instantaneous edges, these Granger causal relations will correspond to relations in the underlying SCM \citep[Theorem 10.3]{peters2017elements}.

\subsection{Hidden Confounding}

Hidden confounders are a critical problem in the time-series context: when they exist,
Granger causality is not guaranteed to correspond to the true causal graph anymore \cite[Theorem 10.3]{peters2017elements}\footnote{For instance, if an unobserved time-series $U$ causes both time-series $X$ and $Y$, then the past of $X$ can help predict the future of $Y$, even though there is no causal link between them.}.
Inspired by proxy-based methods from causal inference (e.g. \citet{louizos2017causal}, see \cref{sec:related}), we present a method for applying ACD to the hidden confounding setting. 
First, we extend the amortized encoder $\qphi$ to predict an additional variable. Then, we encourage this variable to model the hidden confounder by applying a structural bias -- depending on the type of unobserved variable that we want to model, its predicted value is utilized differently by the remaining model.
The decoder remains responsible for modeling the dynamics, and now also processes the predictions for the unobserved variable.
While this setup might not allow us to identify the hidden confounders, and it is still true that the predicted Granger causal relations may not correspond to the underlying SCM, the data-driven approach underlying ACD can benefit our model: 
by training across samples with different underlying causal graphs, our model has access to substantially more information about the causal dynamics at hand and we show empirically that it can learn to mitigate the effects of the hidden confounders. 

We consider two types of hidden confounders which were chosen to cover a wide range of potential confounders as one might encounter in practice. First, we introduce a temperature variable that confounds all observed variables by influencing the strength of their causal relations. This temperature is sampled separately for each sample, and remains constant throughout each sample. This is a realistic example of a global variable that may influence physical dynamics. Second, we introduce a hidden variable that behaves just like the observed variables, i.e. it may affect or be affected by the observed variables through the same causal relations, and its value changes across time. This confounder is inspired by the introductory example where one wants to infer the causal relations between neurons based on their spiking pattern. In this scenario, it is virtually impossible to record all relevant neurons. The resulting unobserved neurons behave and influence the observed neurons in a similar fashion as the unobserved time-series confounder introduced here. In both scenarios, we extend the encoder to predict this hidden variable, and feed that prediction into the decoder. We provide more details in \cref{sec:exp_unobserved}.

\section{Related Work}\label{sec:related}

A range of approaches to causal discovery in both temporal and non-temporal data exist \citep{heinze2018causal,spirtes2000causation,peters2017elements}.
One common class is \textit{constraint-based}, relying on conditional independence testing to uncover an underlying DAG structure or equivalence class \citep{spirtes2000causation}.
These methods predict a single graph $\hat{\mathcal{G}}$ (or equivalence class) for all samples.
There is no notion of fitting a dynamics model for time-series methods in this class \citep{entner2010causal}. 
Another common class of methods for causal discovery is \textit{score-based} \citep{chickering2002optimal, bengio2019meta}. 
Here, a score function $h$ is chosen, and the methods perform a search through graph space to optimize this score, i.e. $\hat{\mathcal{G}} = \textrm{argmin}_{\mathcal{G}} \ h(\mathcal{G})$.
Our proposed decoder-based inference (\cref{eq:tta}) can be seen as score-based causal discovery with a \textit{learned} score function $\mathcal{L} \circ f_{\theta_\star}$.
A third class of methods fits a (possibly regularized) dynamics model $f$ and then analyzes its form to produce a causal graph estimate, by using linear dynamics \citep{JMLR:v11:hyvarinen10a}, recurrent models \citep{tank2018neural,khanna2019economy,nauta2019causal}, or other deep-learning based approaches \citep{wu2020discovering,lachapelle2019gradient,zheng2020learning}.
See \cref{sec:background} for discussion.

The range of approaches to causal discovery relevant to ours include \citet{li2018learning}, who propose to learn a linear mixed effects model across samples; concurrent work explores amortized deep learning of differing types of causal structure \citep{li2020causal,ke2020amortized} and the relationship between GNNs and causal effect inference \citep{zevcevic2021relating}.
Additionally relevant are other approaches for temporal data: works such as \citet{peters2013causal} or \citet{eichler2012causal} use independence or additivity assumptions.
Another adjacent category of causal discovery work explores
the idea of jointly learned causal structure across examples, including in the setting where a number of related datasets are collected \citep{dhir2020integrating,huang2019specific,huang2020causal,shimizu2012joint,tillman2014learning,huang2020causaldiscovery}.

There is little systematic study of hidden confounding in the time-series setting. Some empirical work in the non-temporal domain shows that encoder-based models with enough proxies (variables caused by hidden confounders) can improve causal inference under hidden confounding \citep{louizos2017causal,parbhoo2020information}, and theoretical work proves the identifiability of latent variables from proxies under some assumptions \citep{kruskal1977three, allman2009identifiability}. 

Several works have used graph neural networks \citep{battaglia2016interaction,santoro2017simple,kipf2018neural} or attention mechanisms \citep{vaswani2017attention,fuchs2019end,goyal2019recurrent,van2018relational} to infer relations between time-series. 
\citet{alet2019neural} propose a meta-learning algorithm to additionally model unobserved variables.
While these approaches model object relations in a number of ways, they are not explicitly designed to infer \textit{causal} graphical structure.

The probabilistic implementation of ACD is based on the Neural Relational Inference (NRI) model \citep{kipf2018neural}. We extend this work by new inference methods using test-time adaptation and new algorithms to handle hidden confounders. Moreover, we provide a proof that relates the zero-edge function to Granger causality, which allows for a causal interpretation of the inferred edges. Subsequently, we apply our model to a different problem than NRI, namely (Granger) causal discovery, and show that it outperforms the current state of the art for this type of problem. Last but not least, we show that our model achieves strong causal discovery performance even under noise and hidden confounding, an accomplishment that is -- to the best of our knowledge -- new in this field.

\begin{figure}[t]
    \centering
    \includegraphics[height=10.5em]{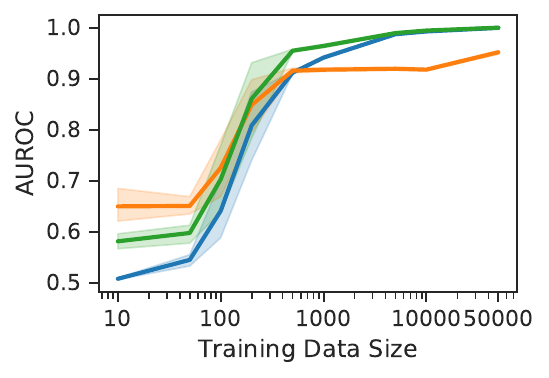}
    \includegraphics[height=10.5em]{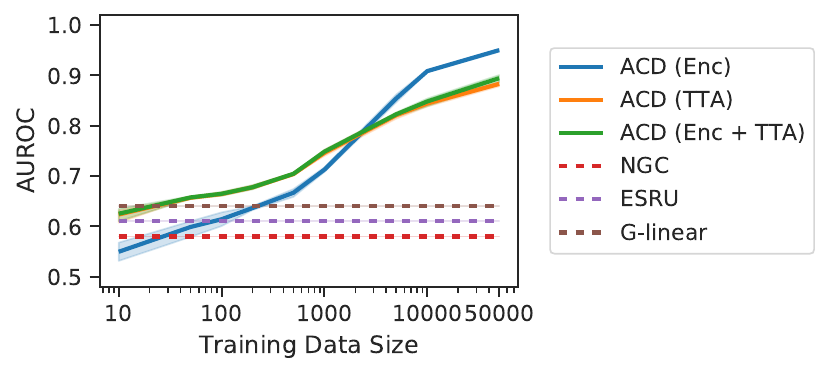} \hfill
    \caption{
    Causal discovery performance (in AUROC) on the particles dataset (A-left) and Kuramoto (B-right).
    ACD improves with more training data, outperforming previous approaches with as few as 50 available training samples on Kuramoto.
    In the high-data regime, encoder inference (\textit{Enc}) is best, while test-time adaptation (\textit{TTA} and \textit{Enc+TTA}) is superior in low-data settings.
    }
    \label{fig:training_samples}
\end{figure}

\section{Experiments}

\paragraph{Implementation} \label{ref:models}
We measure causal discovery performance by the area under the receiver operator curve (AUROC) of predicted edge probabilities over test samples.
We compare to recurrent models (\citet{tank2018neural, khanna2019economy}), a mutual-information (MI) based model by \citet{wu2020discovering} and several baselines implemented by those authors, including MI (unmodified), 
transfer entropy \citep{schreiber2000measuring}, and linear Granger causality. More details in \cref{app:experimental-details}, our code is available at \href{https://github.com/loeweX/AmortizedCausalDiscovery}{\texttt{github.com/loeweX/AmortizedCausalDiscovery}}. 

\subsection{Fully Observed Amortized Causal Discovery} \label{sec:exp_causal}

We test ACD on three datasets: two fully-observed physics simulations (Kuramoto and Particles) and the Netsim dataset of simulated fMRI data \citep{smith2011network}.
Note, in contrast to the physics simulations used in \citet{kipf2018neural}, we generate data with \textit{asymmetric} connectivity matrices to represent causal relations. 

First, we test our method on the \textbf{Kuramoto dataset}, which contains five 1-D time-series of phase-coupled oscillators \citep{kuramoto1975self}.
We find that ACD greatly outperforms all approaches for Granger causal discovery that we compare against (\cref{tab:causal_inference_kuramoto}). In contrast to these approaches, ACD achieves this result \textit{without} fitting to the test samples.

\begin{wraptable}[12]{r}{0.51\textwidth} %
    \caption{AUROC for causal discovery on Kuramoto dataset. 95\% confidence interval shown.} 
    \label{tab:causal_inference_kuramoto}
    \centering
    \small
    \begin{tabular}{cc}
    
        \toprule
        Method & AUROC \\
        \midrule
        MPIR \citep{wu2020discovering}                      & 0.502 $\pm$ 0.006 \\
        Transfer Entropy \citep{schreiber2000measuring}     & 0.560 $\pm$ 0.005 \\
        NGC \citep{tank2018neural}                          & 0.574 $\pm$ 0.018 \\
        eSRU \citep{khanna2019economy}                      & 0.607 $\pm$ 0.001 \\
        Mutual Information                                                  & 0.616 $\pm$ 0.000 \\
        Linear Granger Causality                                            & 0.647 $\pm$ 0.003 \\
        \midrule
        Amortized Causal Discovery                          & \textbf{0.952 $\pm$ 0.003} \\
        \bottomrule
    \end{tabular} 
\end{wraptable}
Additionally, we find that ACD can indeed utilize samples with different underlying causal graphs -- its performance improves steadily with increasing training data size (\cref{fig:training_samples}).
Nonetheless, it is also applicable to the low-data regime: when applying ACD with test-time adaptation (TTA), it requires less than $50$ training samples to outperform all previous approaches.
We note that the baseline performance here is worse than presented elsewhere in the literature -- this is because we do not evaluate the prediction of self-connectivity, which is the easiest connectivity to predict.

In our second experiment, we apply ACD to the \textbf{particles dataset}. This dataset models five particles that move around a two-dimensional space, with some particles influencing others uni-directionally by pulling them with a spring.
Since all previous methods were intended for one-dimensional time-series, we were unable to evaluate them in this domain.
ACD, on the other hand, is readily applicable to higher-dimensional data, and performs almost perfectly on this dataset with \textbf{0.999 AUROC}.

In both experiments, causal relation prediction with the learned encoder (Enc - \cref{eq:enc-optimization}) performs best in the high-data regime, while test-time adaptation (TTA - \cref{eq:tta}) improves the performance in low-data settings (\cref{fig:training_samples}).
This benefit of TTA can be largely attributed to two effects.
First, TTA closes the amortization gap of the encoder \citep{cremer2018inference}. 
Second, TTA overcomes the encoder's overfitting on the training data (as seen in the training curves in \cref{app:training_curves}) by adapting to the individual test samples. %
On the particles dataset, initializing TTA with the encoder's prediction (Enc+TTA) improves over a random initialization (TTA) as the encoder improves; but we do not observe this effect on the Kuramoto dataset.

\begin{wraptable}[12]{R}{0.51\textwidth}
    \vspace{-11pt}	
    \caption{AUROC for causal discovery on Netsim dataset. 95\% confidence interval shown.} 
    \label{tab:causal_inference_netsim}
    \centering
    \small
    \begin{tabular}{cc}
        \toprule
        Method & AUROC \\
        \midrule
        MPIR \citep{wu2020discovering}                      & 0.484 $\pm$ 0.017 \\
        Transfer Entropy \citep{schreiber2000measuring}     & 0.543 $\pm$ 0.003 \\
        NGC \citep{tank2018neural}                          & 0.624 $\pm$ 0.020 \\
        eSRU \citep{khanna2019economy}                      & 0.670 $\pm$ 0.015 \\
        Mutual Information                                  & \textbf{0.728 $\pm$ 0.002} \\
        Linear Granger Causality                            & 0.503 $\pm$ 0.004 \\
        \midrule
        Amortized Causal Discovery                          & \textbf{0.688 $\pm$ 0.051} \\
        \bottomrule
    \end{tabular} 
\end{wraptable}

Finally, we apply ACD to the \textbf{Netsim} dataset \citep{smith2011network} of simulated fMRI data. Here, the task is to infer the underlying connectivity between 15 brain regions across 50 samples. A single graph underlies all samples, allowing us to demonstrate ACD's applicability to the classical setting.
We replace the amortized encoder $\qphi$ with a global latent distribution $\qz$, optimize it through the decoder, and then use test-time adaptation (TTA).
Even though our model cannot benefit from its data-driven design here, it performs comparably to methods that are intended for use in the single-graph setting (\cref{tab:causal_inference_netsim}).

\begin{figure*}[t]
    \centering
    \includegraphics[height=12em]{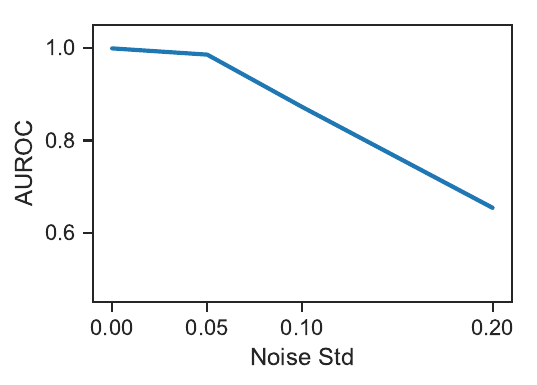}
    \includegraphics[height=12em]{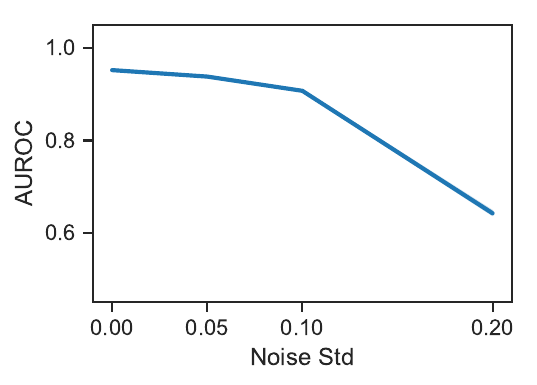} \hfill
    \caption{
    Causal discovery performance (in AUROC) on the particles dataset (A-left) and Kuramoto (B-right) for different levels of observation noise across five seeds.
    For comparison, the standard deviation of both noiseless datasets is about 0.6.
    }
    \label{fig:noisy_performance}
\end{figure*}

\subsubsection{Noisy Data} \label{app:noisy_data}

Handling noisy data is a key challenge in causal discovery.
Here, we show that ACD is robust to a certain amount of observational noise in both the Particles and Kuramoto tasks.

We test ACD's performance for different levels of observation noise added to the Particles and Kuramoto datasets (see \cref{fig:noisy_datasets} in the Appendix for visual examples). 
We sample this additive noise from a zero-mean normal distribution with standard deviations varying between 0.0 and 0.2. For comparison, the input samples in both the Particles and Kuramoto datasets have a standard deviation of 0.6.
For each dataset and each standard deviation, we train across five random seeds. In \cref{fig:noisy_performance}, we show the average AUROC on a test set with the same noise scheme applied. ACD's performance degrades gracefully when stronger observation noise is added. In the highest noise scenario, when the noise is sampled from a normal distribution with a standard deviation of 0.2, the performance of ACD matches the performance of the baselines from \cref{tab:causal_inference_kuramoto} on \textit{noiseless} data.

\begin{figure}[t]
\begin{floatrow}
    \ffigbox[0.48\textwidth]{%
    \includegraphics[height=12em]{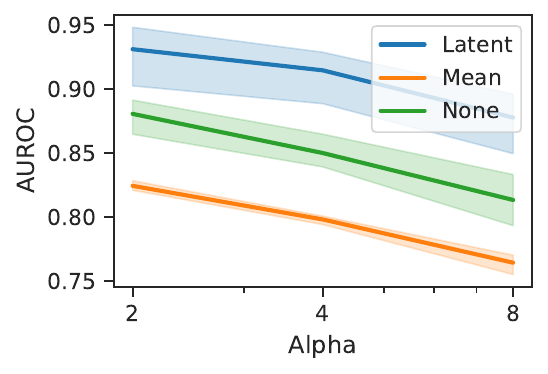}
    }{%
      \caption{
      AUROC with unobserved temperature.
    ACD with a \textit{latent} variable 
    outperforms a baseline which imputes a \textit{mean} temperature, and a learned fixed-temperature decoder (\textit{None}).
      }%
      \label{fig:temperature}
    }\hfill
    \ffigbox[0.48\textwidth]{%
        \includegraphics[height=12em]{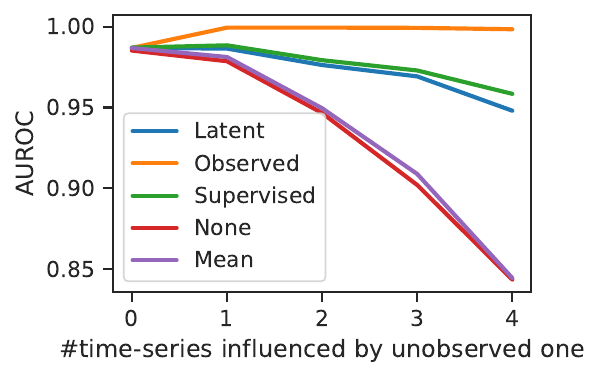}
    }{%
      \caption{
      AUROC with unobserved time-series. As more time-series are influenced by the unobserved one (x-axis), the benefit of using an additional \textit{latent} variable for modeling its effects grows.
      }%
      \label{fig:influenced_AUROC}
    }
\end{floatrow}
\end{figure}

\subsection{Amortized Causal Discovery under Hidden Confounding}\label{sec:exp_unobserved}

\subsubsection{Latent Temperature}
In this experiment, we use the particles dataset and vary an unobserved temperature variable, which modulates how strongly the particles exert force on each other -- higher temperatures result in stronger forces and a more chaotic system.
For each $\xvec_s$, we sample an independent temperature $c \sim \textrm{Categorical}([\frac{\alpha}{2}, \alpha, 2\alpha])$ from a categorical distribution with $\alpha \in \mathds{R}$ and equal probabilities. 
We predict this unobserved temperature by extending the amortized encoder with an additional latent variable, which models a uniform distribution.
Then, we add a KL-Divergence between this posterior and a uniform prior on the interval $[0, 4\alpha]$ to our variational loss. To allow for learning in this setting, we introduce an inductive bias: we use a decoder which matches the true dynamics $g$ given the predicted temperature and causal relations.
See \cref{app:temperature} for more details and additional results.

\paragraph{Results} \cref{fig:temperature} shows the causal discovery results across different values of $\alpha$. ACD enhanced with an additional latent variable (\textit{Latent}) outperforms both tested baselines across all temperatures: \textit{Mean}, which uses the same ground-truth decoder as \textit{Latent} and fixes the decoder temperature to be the mean of the categorical distribution, and \textit{None}, which does not model $c$ explicitly and trains an MLP decoder.
Additionally, this method achieves high predictive performance on the unobserved temperature variable: for $\alpha$ = 2, temperature prediction obtains 0.888 $R^2$, 0.966 AUROC and 0.644 accuracy. 
These results indicate that we can model an unobserved temperature variable, and thus improve robustness under hidden confounding.

\subsubsection{Unobserved Time-Series}

\begin{wrapfigure}[20]{R}{0.51\textwidth}
    \vspace{-35pt}	
    \centering
    \includegraphics[width=0.9 \linewidth]{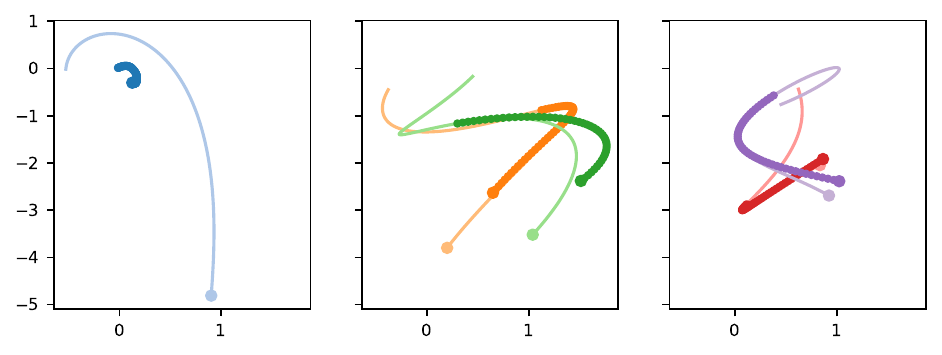} \\
    \includegraphics[width=0.9 \linewidth]{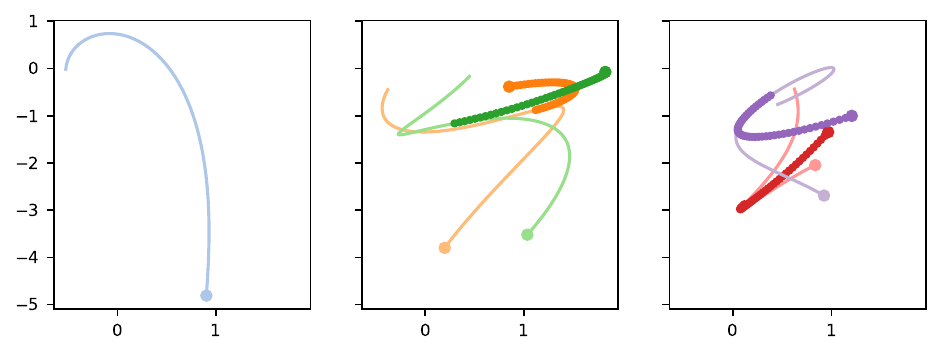}   
    \caption{
    Trajectory prediction with an unobserved time-series (TS).
    Faded: ground truth. Bold: prediction, starts after observing the first half of the ground truth. Dots denote end of TS.
    Top: ACD with \textit{Latent}, bottom: \textit{None} baseline - does not model unobserved TS.
    Left: unobserved TS, middle: TS directly influenced by unobserved, right: remaining TS.
    Though we underestimate the unobserved TS, observed TS prediction improves.
    }%
    \label{fig:unobserved_trajectory}
\end{wrapfigure}

Here, we treat one of the original time-series in the particles dataset as unobserved. It exhibits the same dynamics as the observed time-series, evolving and causally influencing others the same way as before. This challenging setting has received little attention in the literature; \citet{alet2019neural} tackled it with mixed success.
We model the unobserved time-series by extending the amortized encoder with an additional latent variable and applying a suitable structural bias: the latent prediction $\zvec_u^t$ for time-steps $t = \{1,...,T \}$ is treated in the same way as the observed time-series $\xvec$.
Its entire trajectory is used by the encoder to predict causal relations, and its value at the current time-step is fed into the decoder. See \cref{app:unobserved_time} for more details and additional results.

\paragraph{Results}
\cref{fig:influenced_AUROC} shows how the causal discovery AUROC depends on the number of observed time-series directly influenced by the unobserved one. 
When this number is zero, all tested approaches perform the same.
With growing numbers of influenced time-series, the baselines that either ignore the missing time-series (\textit{None}) or impute its value with the average of the observed time-series over time (\textit{Mean}) deteriorate strongly.
In contrast, the proposed ACD with a \textit{Latent} variable stays closer to the performance of the fully \textit{Observed} baseline. As shown in \cref{fig:unobserved_trajectory}, it also improves the future trajectory prediction of the observed time-series. A \textit{Supervised} baseline, that uses the (usually unavailable) ground-truth trajectories to optimize the prediction of the unobserved time-series, improves only slightly over our approach.
These results indicate that ACD can use latent variables to improve robustness to unobserved time-series.

\section{Conclusion}
We introduce ACD, a framework for causal discovery in time-series data which can leverage the information that is shared across samples.
We provide a probabilistic implementation of this framework, and demonstrate significant performance gains over the existing literature when predicting causal relations, both in the fully observed setting and with noise and hidden confounding.

Despite this improvement in performance over previous work, several limitations remain.
Our assumptions from Sec. \ref{sec:method} regarding shared dynamics and edges in the graph, are not verifiable in practice; however, this is standard in causal inference, which frequently relies on untestable assumptions such as ignorability or consistency.%

We conducted all our experiments on simulated data; as a result, they are not particularly realistic: real-world data is more complex and potentially misspecified. 
This limitation is in line with the data used in related works (e.g. \citet{khanna2019economy}), and it remains an exciting direction for future work to improve modeling and experimentation in more realistic settings. 

Finally, our contribution is primarily empirical: we propose and attack a novel version of the causal discovery problem -- where samples have different underlying causal graphs but shared dynamics -- and demonstrate that ACD outperforms prior methods by a large margin in this setting. 
It remains an important future direction of inquiry to understand the conditions under which ACD, as well as other Granger causal discovery approaches that inspired ACD (e.g. \citet{tank2018neural, khanna2019economy}), can be guaranteed to identify the correct causal structures.

\acks{Thanks to Thomas Kipf for helpful discussions and to Sara Magliacane, Marco Federici, Gabriele Bani, Joop Pascha, Patrick Forre, Pascal Esser, Maja Rudolph, Elliot Creager, Taylor Killian, and Renjie Liao for their
feedback on the manuscript. David Madras was supported by an NSERC Alexander Graham Bell
Canada Graduate Scholarship-Doctoral (CGS-D). Resources used in preparing this research were
provided, in part, by the Province of Ontario, the Government of Canada through CIFAR, and
companies sponsoring the Vector Institute (\url{www.vectorinstitute.ai/\#partners)}.}

\bibliography{bibliography}

\clearpage
\appendix

\section{Granger Causality of ACD} \label{app:claim-gc}

Here, we show that when constraining the edge-type $e=0$ to be the zero function, time series $i$ does not Granger cause the model prediction of $j$ in ACD.
In the noiseless setting, at the global optimum of sufficiently expressive model classes, this is equivalent to saying that we recover the true Granger causal relations.

\paragraph{Claim.}
If $z_{ij,0} = 1$, $i$ does not Granger cause the model prediction of $j$ in ACD.
\paragraph{Proof.}
According to \cref{def:nonlinear_granger}, time-series $i$ does not cause $j$, if $g_j$ is invariant to $\xvec^{\leq t}_i$. In our model, the decoder represents this non-linear model $g_j$
and consists of two functions. First, it propagates information across edges using \cref{eq:edges}. This function returns a value of zero, if $z_{ij,0} = 1$. This output is used for the second function, described in \cref{eq:nodes}, which does not introduce any new terms that depend on $i$. Thus, if $z_{ij,0} = 1$, the decoder's prediction for $j$ is invariant to $\xvec^{\leq t}_i$, and $i$ does not Granger cause these predictions.

\section{Fully Observed Amortized Causal Discovery} \label{app:experimental-details}

\subsection{Experimental Details}

\subsubsection{Datasets}
\paragraph{Physics Simulations}
To generate these simulations, we follow the description of the underlying physics of \citet{kipf2018neural} for the phase-coupled oscillators (Kuramoto) \citep{kuramoto1975self} and the particles connected by springs. In contrast to their simulations, however, we allow the connectivity matrix, which describes which time-series influences another, to be asymmetric. This way, it describes causal relations instead of correlations.

\begin{figure*}[t]
    \centering
    \includegraphics[height=11em]{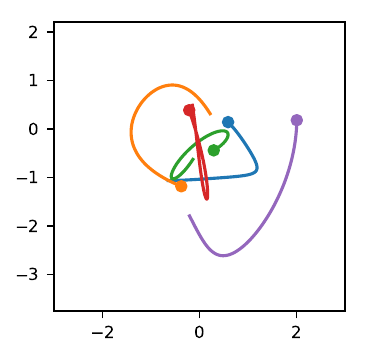}
    \hspace{0.5em}
    \includegraphics[height=11em]{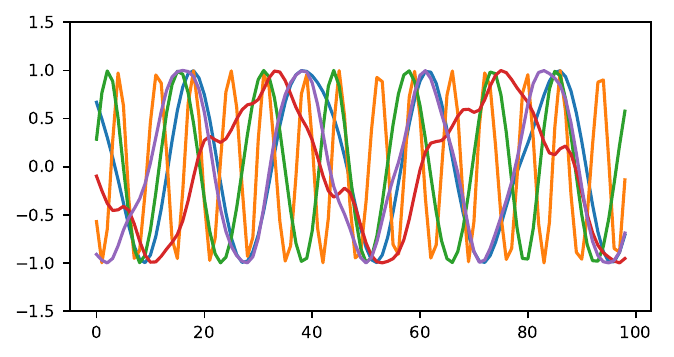} \\
    \vspace{0.5em}
    
    \includegraphics[height=11em]{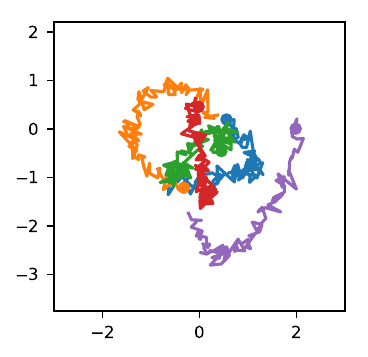}
    \hspace{0.5em}
    \includegraphics[height=11em]{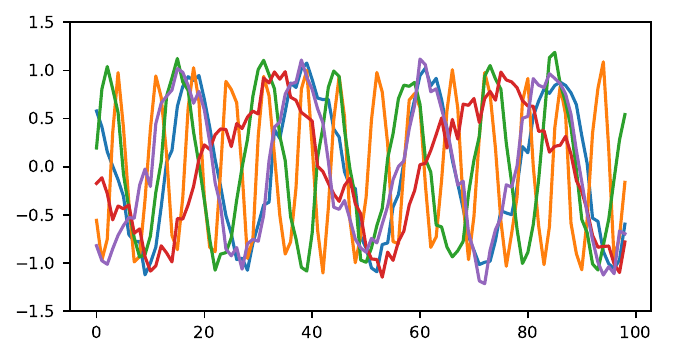} \\
    \vspace{0.5em}
    
    \includegraphics[height=11em]{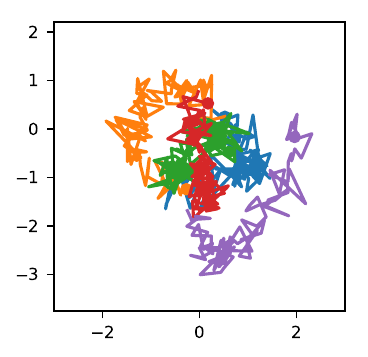}   
    \hspace{0.5em}
    \includegraphics[height=11.6em]{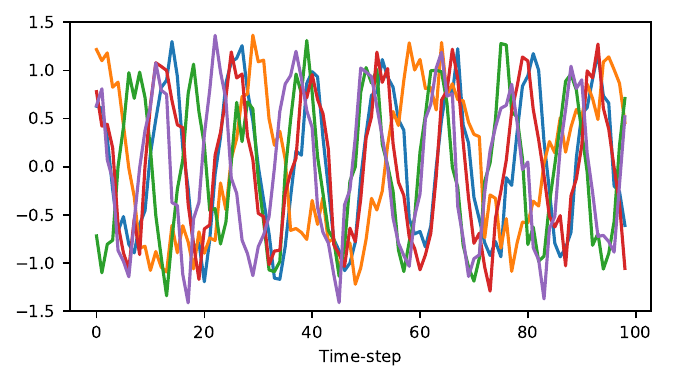}
    \caption{Example trajectories with different levels of observation noise for the Particles dataset (left) and Kuramoto (right). The observation noise is sampled from a zero-mean Gaussian distribution with standard deviation of (from top to bottom) 0.0, 0.1 and 0.2, respectively.}
    \label{fig:noisy_datasets}
\end{figure*}

For both datasets, we generate 50,000 training and 10,000 validation samples. We restrict the number of test samples to 200, since the previous methods we compare to must be refit for each individual sample. We simulate systems with $N=5$ time-series. Our training and validation samples consist of $T=49$ time-steps, while the test-samples are $T=99$ time-steps long. This increased length allows us to infer causal relations on the first half of the data, and to test the future prediction performance on the second half (with $k=\{1,...,49\}$). In Figure \ref{fig:noisy_datasets}, we show some examples of both the particles and Kuramoto datasets in the noiseless setting and with added noise.

\paragraph{Netsim}
The Netsim dataset simulates blood-oxygen-level-dependent (BOLD) imaging data across different regions within the human brain and is described in \citet{smith2011network}. The task is to infer the directed connections, i.e. causal relations, between different brain areas.

The Netsim dataset includes simulations with different numbers of brain regions and different underlying connectivity matrices. In our experiments, we use the data from the third simulation Sim-3.mat as provided by \citet{khanna2019economy}. It consists of samples from 50 subjects, each with the same underlying causal graph, each of length $T=200$ and including $N=15$ different brain regions. Note, that we report worse results than \citet{khanna2019economy}, since we assume self-connectivity for all time-series and only evaluate the causal discovery performance between \textit{different} time-series.

The dataset is very small (50 samples) and due to this, we do not use a training/validation/test split, but use the same 50 points at each phase instead.
While this is not standard machine learning practice, it still facilitates a fair comparison to the other methods, each of which is fit to individual test points.
The purpose of including experiments on this dataset is not to demonstrate generalization ability, but rather to show that our method is flexible enough to work reasonably well even in the classical causal discovery setting (with one shared causal graph, and fitting the model on the test set).

\subsubsection{Architecture and Hyperparameters}

Our model is implemented in PyTorch \citep{paszke2019pytorch}. We did no hyperparameter optimization for model training, but used the settings as described for the NRI model \citep{kipf2018neural}. The latent dimension throughout the model is set to size 256. We optimize our model using ADAM \citep{kingma2014adam} with a learning rate of 0.0005. In the experiments on the particles dataset, the learning rate is decayed by a factor of 0.5 every 200 epochs. We set our batch-size to 128 and train for 500 epochs. The temperature of the Gumbel-Softmax is set to $\tau = 0.5$. During testing, this concrete distribution is replaced by a categorical distribution to obtain discrete edge predictions.

There was no thorough hyperparameter optimization done for test-time adaptation (TTA). Since there was no pre-existing implementation, some hand-tuning was performed.
We use a learning rate of 0.1 for the Kuramoto and particles datasets and 0.01 for Netsim. For each, we run 1000 iterations.

\paragraph{Encoder}
In our experiments, the amortized encoder applies a graph neural network $f_{enc,\phi}$ on the input. It implements two edge-propagation steps along the causal graph:
\begin{align}
    \psivec_j^1 &= f_{\text{emb}}(\xvec_j) \\
    \psivec_{ij}^1 &= f_e^1([\psivec_i^1, \psivec_j^1]) \\
    \psivec_j^2 &= f_v^2(\sum_{i \neq j} \psivec_{ij}^1) \\
    \psivec_{ij} &= f_e^2([\psivec_i^2, \psivec_j^2]) ~~~.
\end{align}
$f_e^1, f_e^2$ and $f_v^2$ are fully-connected networks (MLPs). On both the particles dataset and Netsim, $f_{\text{emb}}$ is an MLP as well (MLP Encoder); on the Kuramoto dataset, we use a 1D CNN with attentive pooling \citep{lin2017structured} instead (CNN Encoder).

When conducting test-time adaptation as described in \cref{eq:tta}, we remove the encoder and model a distribution over $\mathds{G}$ using a non-amortized variational distribution $\qz$ with its initial values sampled from a unit Gaussian.

\paragraph{Decoder}
The decoder implements a single edge-propagation step according to equations 12-14. It uses MLPs for both $f_e$ and $f_v$. To improve performance, we train the decoder to predict several time-steps into the future. For this, we replace the true input $\xvec^t$ with the predicted $\muvec^t$ for $k=10$ steps.

Following our causal formulation of the NRI model, we implement \cref{eq:edges} by masking out the values of the corresponding edges. Thus, the ordering of the edge types is not arbitrary in our setting. 

We note that while this implementation of the decoder assumes a full-time graph of Markov order 1, the full ACD framework does not, and could be implemented using a recurrent architecture to remove this assumption.

Since our physics simulations are differentiable, we can replace the decoder with the ground-truth dynamics and backpropagate through them. We call this setup the simulation decoder.

\paragraph{Variance}
When we report the variance on the ACD results, we collected these across five different random seeds.
Baselines in Kuramoto/Netsim use three seeds each, except for NGC, which uses only one due to a longer runtime (the confidence intervals shown for NGC are confidence intervals on the AUROC itself, whereas all other confidence intervals are based on variance of AUROC across seeds).

\subsubsection{Baselines}

We compare ACD against several baselines:

\paragraph{Neural Granger Causality} From \citet{tank2018neural}, we optimized an MLP or LSTM to do next step prediction on a sample.
We found that the MLP worked best.
The causal links are wherever an input weight is non-zero.
We used ADAM and then line search to find exact zeros.
In this method, we calculate AUROC by running with a range of sparsity hyperparameters ($\lambda = [0, 0.1, 0.2, 0.4, 0.8]$ for Kuramoto and  $\lambda = [0, 0.1, 0.15, 0.2, 0.3, 0.4, 0.5, 0.6, 1]$ for Netsim).
As in \citet{tank2018neural}, we calculate a score $s$ for each edge, where $s = \min \{ \lambda : z_{ij,0} = 1 \}$, and use that score to calculate AUROC.
Code was used from \url{https://github.com/icc2115/Neural-GC}.

\paragraph{ESRU} \citet{khanna2019economy} take a similar approach to \citet{tank2018neural}, but they use economy statistical recurrent units (eSRU), instead of LSTMs.
We found one layer worked best, and used their hyperparameters otherwise.
We use sparsity hyperparameters $[0.1, 0.2, 0.3, 0.4, 0.5]$ for Kuramoto, and $[0, 0.1, 0.15, 0.2, 0.3, 0.4, 0.5, 0.6, 1]$ for Netsim.
Code was used from \url{https://github.com/sakhanna/SRU_for_GCI}.

\paragraph{MPIR} \citet{wu2020discovering} determine where causal links exist by examining the predictive performance change when noise is added to an input variable.
Code for this method and the baselines below was used from \url{https://github.com/tailintalent/causal}.

\paragraph{Transfer Entropy} \citet{schreiber2000measuring} suggest this entropy-like measure between two variables to produce a metric which is likely to be higher when a causal connection exists. We use the implementation by \citet{wu2020discovering}.

\paragraph{Mutual Information} Using the implementation by \citet{wu2020discovering}, we calculate the mutual information between every pair of time series.

\paragraph{Linear Granger Causality} Using the implementation by \citet{wu2020discovering}, this is a linear version of Granger causality where non-zero linear weights are taken as greater causal importance.

We did not run the baselines on the particles dataset since it is two-dimensional and most baselines did not provide an obvious way for handling multi-dimensional time series.
When training ACD on the particles and Kuramoto datasets, we additionally input the velocity (and phase for Kuramoto) of the time-series. Since our chosen NRI encoders and decoders are not recurrent we cannot recover this information in any other way in this model.
This enables a more fair comparison to the recurrent methods, which are able to aggregate this information over several time steps.

\subsection{Additional Experimental Result - Training Curves} \label{app:training_curves}

\begin{figure*}[ht]
    \centering
    \includegraphics[height=12em]{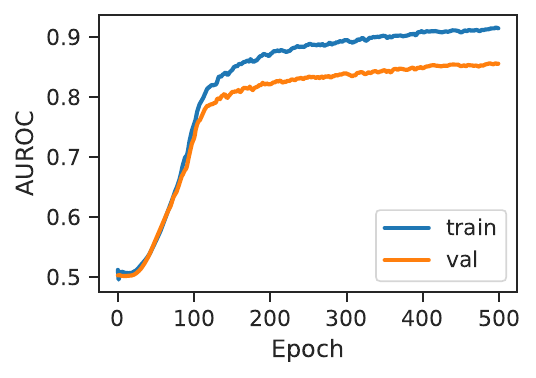}
    \includegraphics[height=12em]{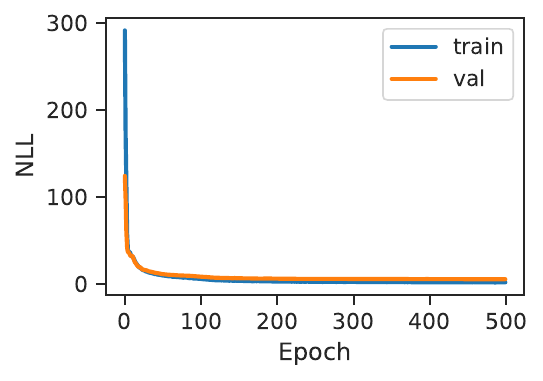}
    \caption{Training curves when training on 100 samples of the particles dataset. The encoder performance (AUROC - left) shows stronger signs of overfitting than the decoder performance (NLL - right).}
    \label{fig:training_curves}
\end{figure*}

\cref{fig:training_curves} shows the training curves when training on 100 training samples of the particles dataset. We observe that the encoder overfits on the training samples, as indicated by the AUROC performance. In contrast, the decoder shows less overfitting as indicated by the negative log-likelihood (NLL) performance.

\section{Amortized Causal Discovery with Unobserved Variables}
\subsection{Temperature Experiments} \label{app:temperature}

\paragraph{Implementation Details}
In this experiment, we use the CNN encoder and a simulation decoder matching the true generative ODE process. Our optimization scheme is the same as before.

For modeling the latent temperature, we output a uniform distribution as our posterior $q_{\phi_c} (c | \xvec)$.
One tricky aspect about this is the KL-Divergence:
\begin{align}
    KL(q_{\phi_c} (c | \xvec) || p(c)) &= - \int q_{\phi_c} (c | \xvec) \log \frac{q_{\phi_c} (c | \xvec)}{p(c)} d\zvec ~~~.
\end{align}
We must ensure that our posterior support is a subset of our prior support. Otherwise, the KL-Divergence is undefined and optimization impossible.
Recall that our prior is a uniform distribution over $[0, 4\alpha]$.

We output two latent parameters $a, b \in \mathds{R}$ for each input and use these values to parametrize a mean $m$ and a half-width $w$ for the uniform distribution.
First, we bound these values to represent a uniform distribution $u_1$ in $[0, 1]$. To achieve this, we let $m_1 = \sigma(a)$ and $w_1 = \sigma(b) * \min{(m_1, 1 - m_1)}$ with $\sigma(x) = \frac{1}{1 + \exp{(-x)}}$.
We then sample a temperature $\hat{c}_1 \sim u_1 = U(m_1 - w_1, m_1 + w_1)$, which is guaranteed to be bounded within $[0, 1]$.
Stopping gradients, we use this temperature sample in the encoder $q_{\phi}(\zvec|\xvec, c)$ to improve the causal discovery performance.

Next, we scale this result to the desired interval $[0, 4\alpha]$. To achieve this, we feed the scaled temperature $\hat{c} = 4 \alpha \hat{c}_1$ into the decoder, and use the scaled distribution $u = U(4 \alpha m_1 - 4 \alpha w_1, 4 \alpha m_1 + 4 \alpha w_1)$ to find our KL term.
We allow gradients to flow through the temperature sample in both the decoder and the distribution in the KL term, which informs our parameter updates.

\begin{figure}
\begin{floatrow}
    \ffigbox[0.45\textwidth]{%
        \includegraphics[height=12em]{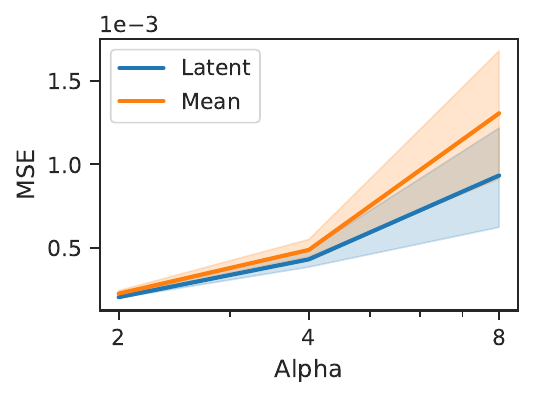}     
    }{%
      \vspace{-1em}
      \caption{MSE (lower better) averaged across 5 random seeds for hidden temperature experiment. MSE for \textit{None} baseline was much worse with MSE $= 0.009, 0.02, 0.04$ for $\alpha = 2, 4, 8$ (not shown in plot).
      }%
      \label{fig:temperature_MSE}
    }\hfill
    \ffigbox[0.45\textwidth]{%
        \includegraphics[height=11em]{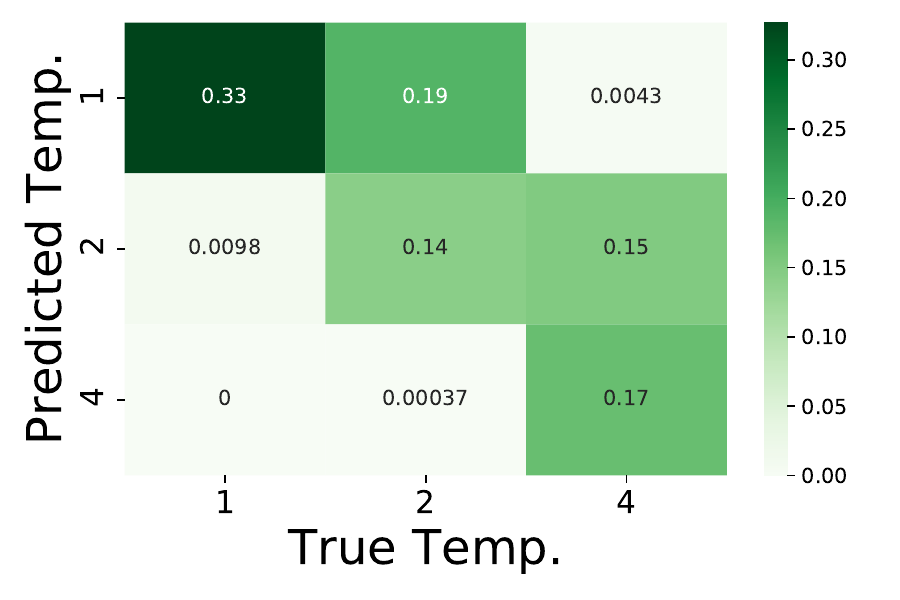}
    }{%
      \vspace{-0.5em}
      \caption{
      Confusion matrix for latent temperature prediction with $\alpha=2$. ACD with Latent's prediction tends to be conservative: it is more likely to predict a too low temperature than a too high one.
      }%
        \label{fig:temperature-prediction}
    }
\end{floatrow}
\end{figure}

\paragraph{Additional Results}
Similarly to \cref{fig:influenced_AUROC}, we show the future prediction performance in MSE across different values of $\alpha$ in \cref{fig:temperature_MSE}. Again we find a slight improvement in performance when using ACD with \textit{Latent} compared to the baselines, although this is a noisier indicator.

Additionally, we evaluate how well the introduced latent variable learns to predict the unobserved temperature. To do so, we use the mean of the predicted posterior uniform distribution.
When a discrete categorical prediction is needed for evaluation, we quantize our results into three bins based on their distance in log-space.
To calculate AUROC in this three category ordinal problem, we average the AUROC between the two binary problems: category 1 vs not category 1, and category 3 vs not category 3 (category 2 vs not 2 is not a valid regression task for the purposes of AUROC which is concerned with ordering, since it is the middle temperature and hence the labels would not be linearly separable).

The confusion matrix between true and predicted temperature in \cref{fig:temperature-prediction} indicates that ACD with Latent's prediction tends to be conservative: it is more likely to predict a too low temperature than a too high one.
This is probably due to higher temperatures incurring larger MSEs, since higher temperature systems are more chaotic and thus less predictable.

\cref{tab:temperature} lists the temperature prediction results across all tested values of $\alpha$.
We find that we can predict the unobserved temperature quite well, especially with respect to ordering (as measured by correlation and AUROC).

\begin{table}[ht]
    \caption{
    Latent Temperature Prediction Metrics. We treat the mean of the outputted interval of the uniform posterior as the predicted temperature.
    For accuracy, this value discretized in log space to get a ternary prediction.
    \textit{AUC (1vAll)} averages the two one-vs-all AUC values, which can be calculated in a 3-category ordinal problem.
    } 
    \label{tab:temperature}
    \centering
    \begin{tabular}{cccc}
        \hline
                    &       & $\alpha$ & \\
                    & 2     & 4     & 8     \\ \hline
        Correlation & 0.888 & 0.844 & 0.661 \\
        Accuracy    & 0.644 & 0.384 & 0.346 \\
        AUROC (1vAll) & 0.966 & 0.935 & 0.843 \\ \hline
    \end{tabular}
\end{table}

\subsection{Unobserved Time-series}\label{app:unobserved_time}

\paragraph{Implementation Details}
For modeling the unobserved time-series, we employ a two-layered, bi-directional long short-term memory (LSTM) \citep{hochreiter1997long} with a latent dimension of size 256.

\paragraph{Additional Results}
The full evaluation of our experiments with an unobserved time-series can be found in \cref{tab:unobserved_particle}. Our results indicate that our proposed method \textit{ACD with latent} predicts the trajectory of the unobserved time-series (unobserved MSE) more accurately than the \textit{Mean} imputation baseline. Even though this prediction is worse than for the \textit{Supervised} baseline, ACD with \textit{Latent} manages to recover the performance of the fully \textit{Observed} baseline better than the \textit{None} and the \textit{Mean} imputation baselines.

\begin{table}[ht]
  \centering
  \caption{Experiments with an unobserved time-series.} 
  \label{tab:unobserved_particle}
    \small
    \begin{tabular}{ccccc}%
    \toprule
    Method & AUROC & Accuracy & MSE & unobserved MSE \\
    \midrule
        Observed & (0.99) & (0.993) & (0.00301) &  - \\
        Supervised & 0.982 & 0.931 & 0.00822 & 0.0164 \\
        None & 0.946 & 0.882 & 0.0119 &  - \\
        Mean & 0.951 & 0.881 & 0.0106 & 0.0397 \\
        \midrule
        ACD with latent & 0.979 & 0.918 & 0.00747 & 0.0375 \\
    \bottomrule
    \end{tabular} 
\end{table}

\begin{figure*}[ht]
    \centering
    \includegraphics[height=9.6em]{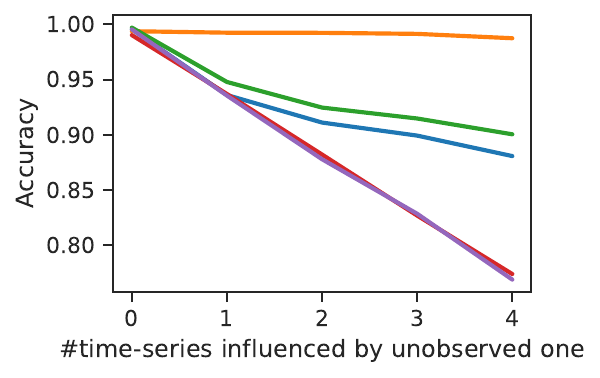}
    \includegraphics[height=9.6em]{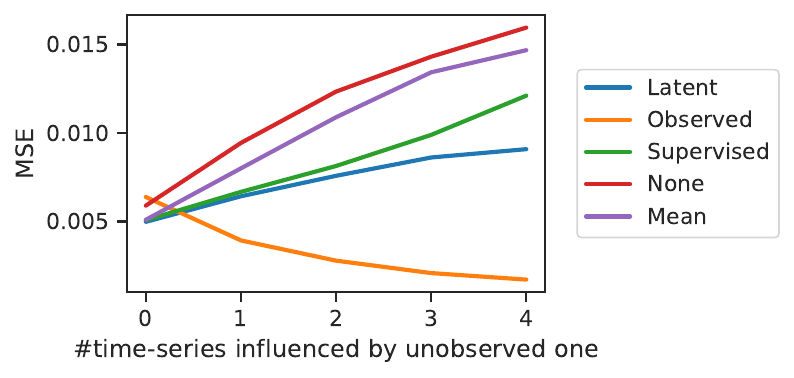} \\
    \caption{Experiments with an unobserved particle. Performance of the various methods depends strongly on how many observed particles are influenced by the unobserved one (x-axis). The more particles that are influenced by the unobserved particle, the stronger the benefit of using an additional \textit{Latent} variable for modeling its effects. Left - causal relation prediction accuracy (higher = better), right - MSE (lower = better).}
    \label{fig:influenced_performance}
\end{figure*}

\begin{figure*}[ht]
    \centering
    \rotatebox[x=0mm,y=6mm]{90}{Observed}
    \hspace*{-1.2mm}
    \includegraphics[width=0.48 \linewidth]{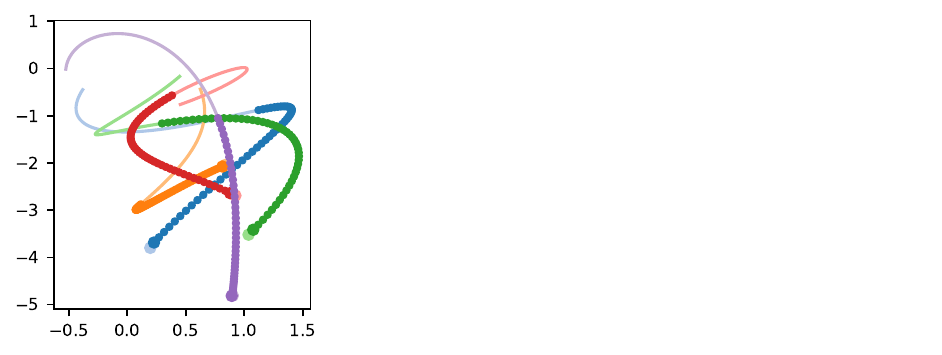} \hfill
    \includegraphics[width=0.48 \linewidth]{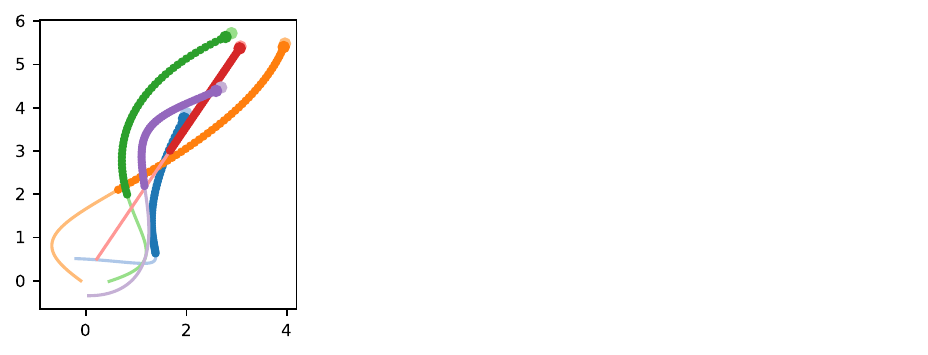} \\
    \rotatebox[x=0mm,y=6mm]{90}{Supervised}
    \hspace*{-1.8mm}
    \includegraphics[width=0.48 \linewidth]{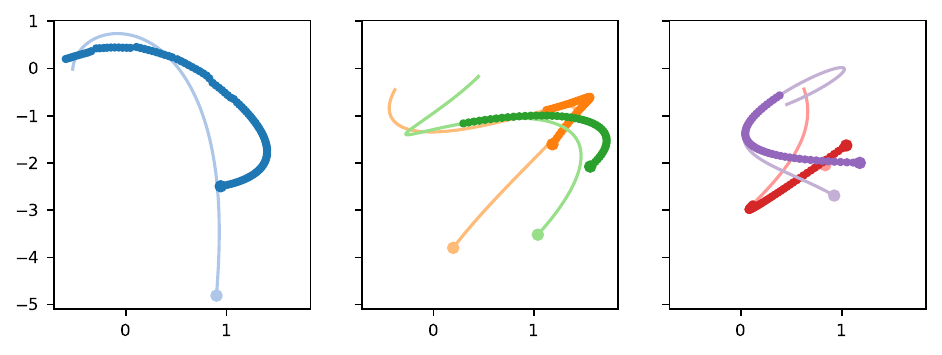} \hfill
    \includegraphics[width=0.48 \linewidth]{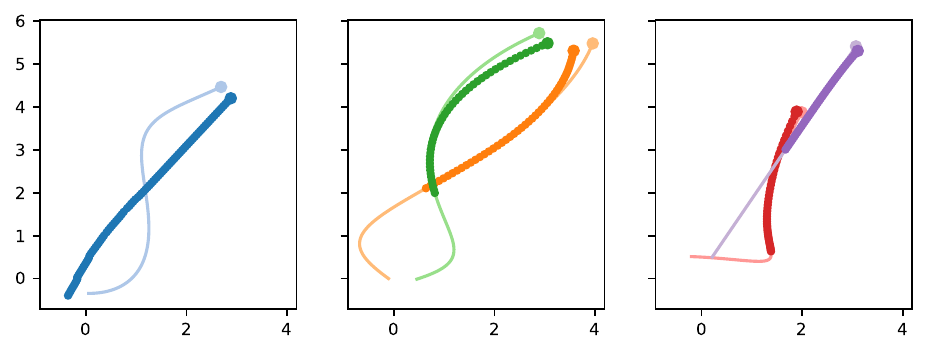} \\
    \rotatebox[x=0mm,y=9mm]{90}{None}
    \includegraphics[width=0.48 \linewidth]{figs/springs5_unobserved1_nomodel/springs_0.pdf} \hfill
    \includegraphics[width=0.48 \linewidth]{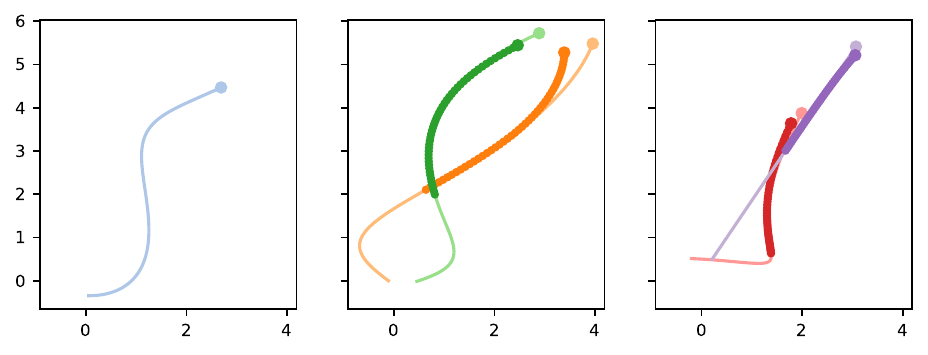} \\
    \rotatebox[x=0mm,y=9mm]{90}{Mean}
    \includegraphics[width=0.48 \linewidth]{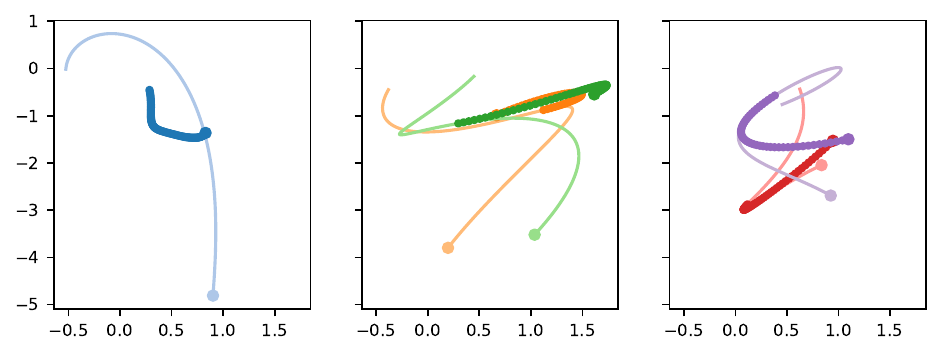} \hfill
    \includegraphics[width=0.48 \linewidth]{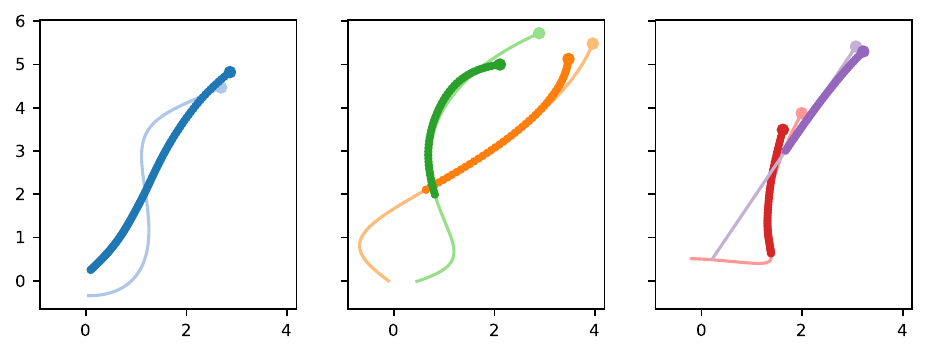} \\
    \rotatebox[x=0mm,y=9mm]{90}{Latent}
    \includegraphics[width=0.48 \linewidth]{figs/springs5_unobserved1/springs_0.pdf} \hfill
    \includegraphics[width=0.48 \linewidth]{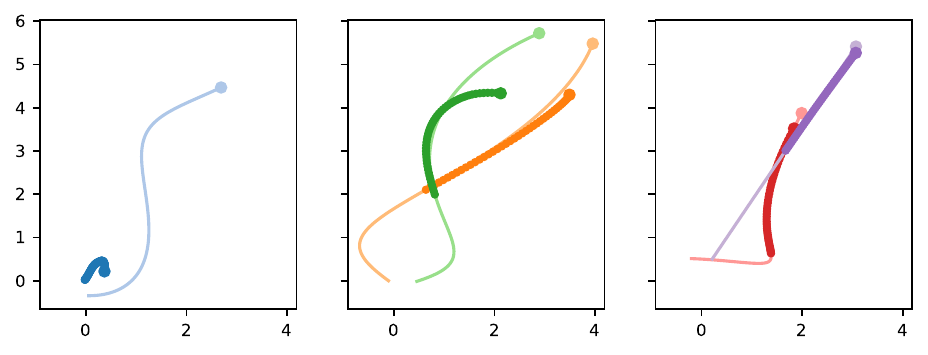} \\
    \caption{Predicted trajectories for all tested methods in the unobserved time-series experiment for two samples (left/right). From top to bottom: Baselines -- observed, supervised, none and mean; proposed ACD with latent. The faded lines depict the ground truth trajectory; bold lines are the trajectories predicted by the model and they start after initializing the model using first half of the ground truth. Dots denote the end of the trajectories. Except for the fully observed baseline, the first panel shows the ground truth and prediction for the unobserved time-series. The second panel shows the trajectories of all time-series that are directly influenced by the unobserved one. The third panel shows the trajectories of all time-series that are not directly influenced by the unobserved one.}
    \label{fig:springs5}
\end{figure*}

\cref{fig:influenced_performance} shows the performance of the tested methods dependent on the number of time-series that are influenced by the unobserved one. In addition to \cref{fig:influenced_AUROC} in our Experiments section, these plots show the achieved accuracy and MSE results. The general trends are the same. \cref{fig:springs5} shows example trajectories and the corresponding predictions for all tested methods.

\begin{figure}[ht] 
    \centering
    \rotatebox[x=0mm,y=12mm]{90}{Observed}
    \hspace*{0.2mm}
    \includegraphics[width=0.4 \linewidth]{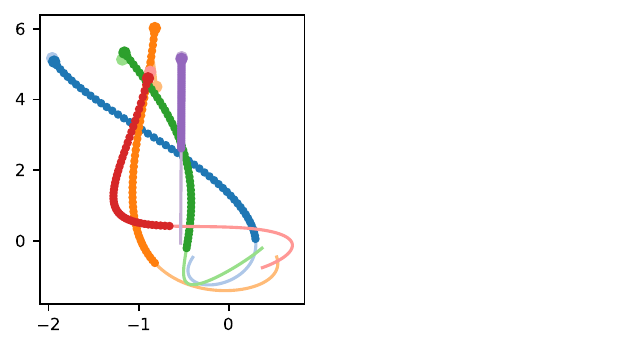} \hfill
    \includegraphics[width=0.4 \linewidth]{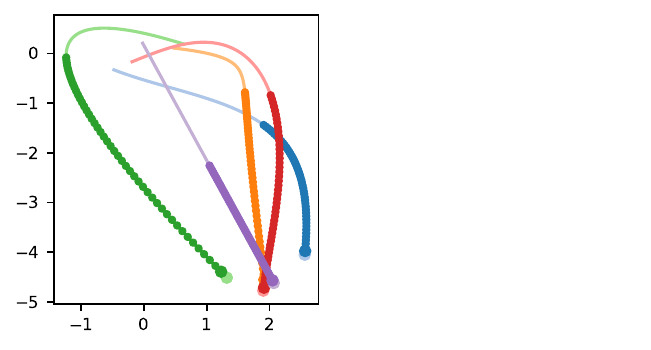} \\
    \rotatebox[x=0mm,y=11mm]{90}{Supervised}
    \hspace*{-1.8mm}
    \includegraphics[width=0.4 \linewidth]{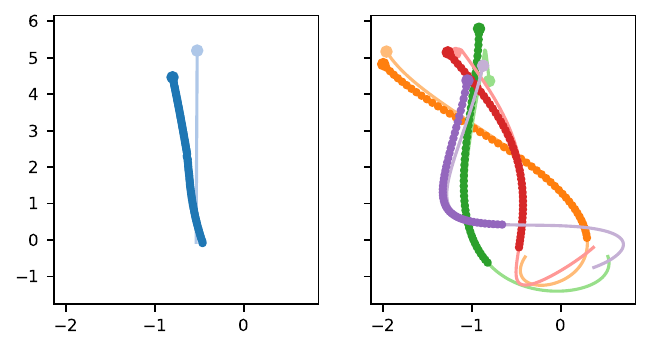} \hfill
    \includegraphics[width=0.4 \linewidth]{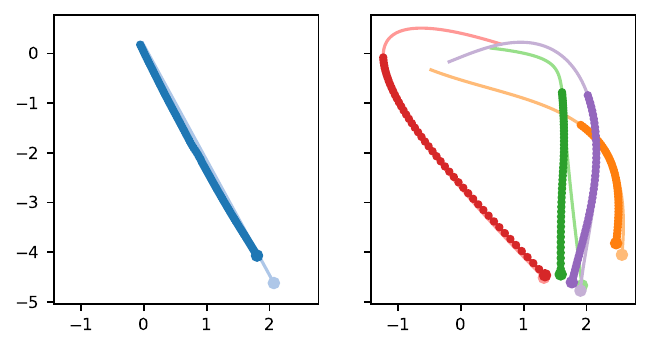}\\
    \rotatebox[x=0mm,y=15mm]{90}{None}
    \includegraphics[width=0.4 \linewidth]{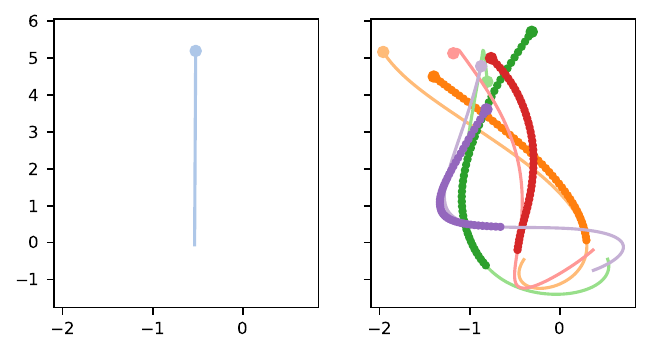} \hfill
    \includegraphics[width=0.4 \linewidth]{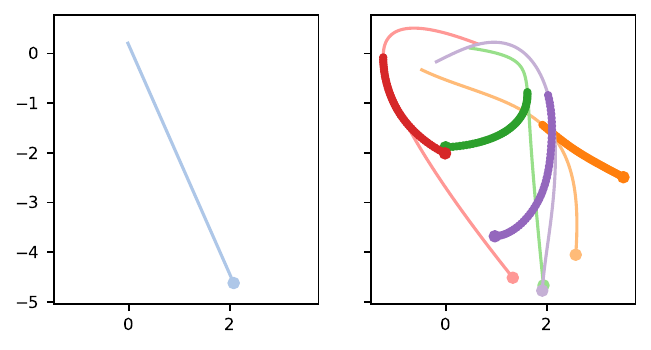}\\
    \rotatebox[x=0mm,y=15mm]{90}{Mean}
    \includegraphics[width=0.4 \linewidth]{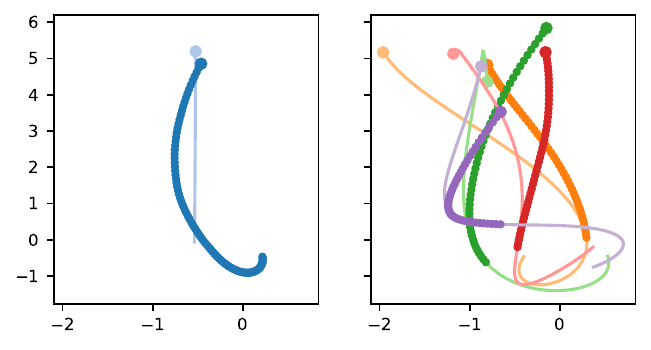} \hfill
    \includegraphics[width=0.4 \linewidth]{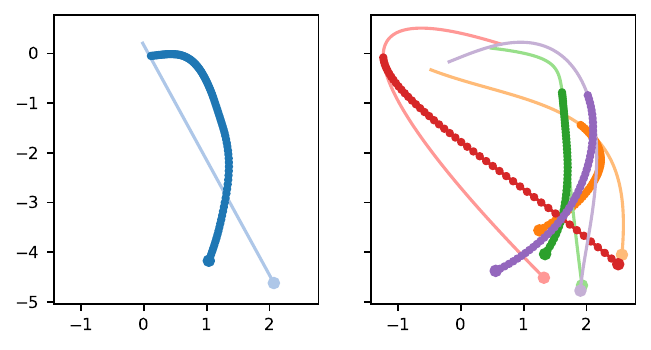} \\
    \rotatebox[x=0mm,y=15mm]{90}{Latent}
    \includegraphics[width=0.4 \linewidth]{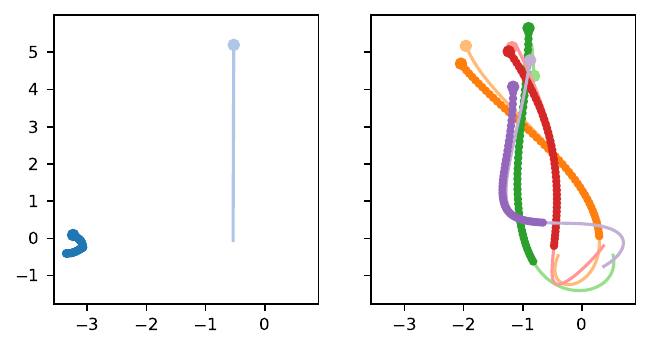} \hfill
    \includegraphics[width=0.4 \linewidth]{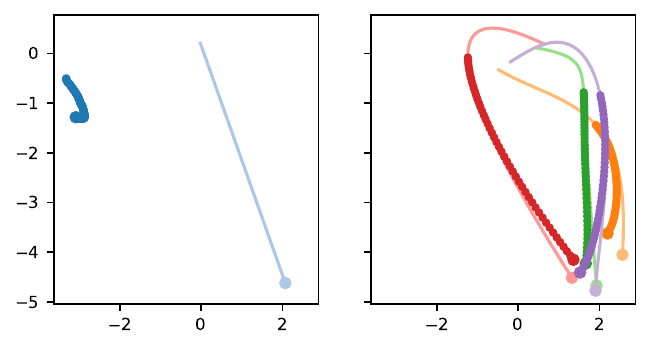} \\
    \caption{
    Predicted trajectories for all tested methods when the unobserved time-series influences all observed ones, but stay uninfluenced itself for two samples (left/right). From top to bottom: Baselines -- observed, supervised, none and mean; proposed ACD with latent. The faded lines depict the ground truth trajectory; bold lines are the trajectories predicted by the model, and they start after initializing the model using the first half of the ground truth. Dots denote the end of the trajectories. Except for the fully observed baseline, the first panel shows the ground truth and prediction for the unobserved time-series. The second panel shows the trajectories of all observed time-series (which are all influenced by the unobserved one).}
    \label{fig:springs5_uninfluenced_influencer}
\end{figure}

\paragraph{Additional Experiment: Uninfluenced Influencer}
Predicting the trajectory of a time-series that influences only a small number of observed time-series and is (invisibly) influenced by them is arguably very difficult. In this follow-up experiment, we reduce the difficulty of this problem by adding two assumptions: (1) the unobserved time-series influences \textit{all} observed time-series and (2) it is not influenced by any of the observed time-series. This way, we gain more information about its trajectory (due to (1)) and its trajectory becomes easier to predict (due to (2)). Indeed, in this setup, \textit{ACD with latent} manages to almost completely recover the performance of the fully observed baseline (\cref{tab:unobserved_uninfluenced_influencer}). In contrast, the performance of the \textit{None} and \textit{Mean} imputation baselines worsens considerably in this setting. Now, all time-series are influenced by the unobserved particle -- making their prediction harder when not taking into account this hidden confounder. \cref{fig:springs5_uninfluenced_influencer} shows example trajectories and the corresponding predictions for all tested methods in this setting.

\begin{table}[ht]
  \centering
  \caption{Experiments with an unobserved time-series that influences all observed time-series, but is not influenced by them.} 
  \label{tab:unobserved_uninfluenced_influencer}
    \small
    \begin{tabular}{ccccc} %
    \toprule
    Method & AUROC & Acuracy & MSE & unobserved MSE \\
    \midrule
        Observed & (1.0) & (0.997) & (0.0193) &  - \\
        Supervised & 1.0 & 0.993 & 0.024 & 0.000615 \\
        None & 0.829 & 0.76 & 0.0431 &  - \\
        Mean & 0.853 & 0.782 & 0.0365 & 0.0357 \\
        \midrule
        ACD with latent & 1.0 & 0.994 & 0.0251 & 0.137 \\
    \bottomrule
    \end{tabular} 
\end{table} 

\end{document}